\documentclass[10pt,journal,compsoc]{IEEEtran}

\usepackage{cite}
\usepackage{amsmath,amssymb,amsfonts}
\usepackage{algorithmic}
\usepackage{graphicx}
\usepackage{subcaption}
\usepackage{makecell}
\usepackage{hyperref}
\renewcommand{\vec}[1]{\mathbf{#1}}
\newcommand{\mat}[1]{\mathbf{#1}}
\newcommand{\set}[1]{\mathcal{#1}}

\begin{document}
\title{EmbedTrack -- Simultaneous Cell Segmentation and Tracking Through Learning Offsets and Clustering Bandwidths}
\author{Katharina~Löffler and Ralf~Mikut
\thanks{We are grateful for funding by the Helmholtz Association in the program Natural, Artificial and Cognitive Information Processing (RM) and HIDSS4Health - the Helmholtz Information \& Data Science School for Health (KL, RM). The funders had no role in study design, data collection and analysis, decision to publish, or preparation of the manuscript.}
\thanks{
Katharina Löffler is with the Institute for Automation and Applied Informatics and with the Institute of Biological and Chemical Systems - Biological Information Processing, Karlsruhe Institute of Technology, 76344 Eggenstein-Leopoldshafen, Germany (email: katharina.loeffler@kit.edu).}
\thanks{
Ralf Mikut is with the Institute for Automation and Applied Informatics, Karlsruhe Institute of Technology, 76344 Eggenstein-Leopoldshafen, Germany (email: ralf.mikut@kit.edu).}}

\maketitle

\begin{abstract}
A systematic analysis of the cell behavior requires automated approaches for cell segmentation and tracking. While deep learning has been successfully applied for the task of cell segmentation, there are few approaches for simultaneous cell segmentation and tracking using deep learning. Here, we present EmbedTrack, a single convolutional neural network for simultaneous cell segmentation and tracking which predicts easy to interpret embeddings. As embeddings, offsets of cell pixels to their cell center and bandwidths are learned. We benchmark our approach on nine 2D data sets from the Cell Tracking Challenge, where our approach performs on seven out of nine data sets within the top 3 contestants including three top 1 performances. The source code is publicly available at \href{https://git.scc.kit.edu/kit-loe-ge/embedtrack}{https://git.scc.kit.edu/kit-loe-ge/embedtrack}.
\end{abstract}


\section{Introduction}\label{sec:introduction}
The directed movement of cells is vital for numerous biological processes such as wound healing, embryonic development or immune response. However, to perform a quantitative analysis of the cell behavior, hundreds or thousands of cells need to be tracked over time. Therefore, automated cell tracking approaches are needed.

 For long, cell tracking has been dominated by traditional approaches such as simple Nearest Neighbors linking~\cite{ulman_objective_2017, stegmaier_fuzzy-based_2017}, Bayesian filters~\cite{chang_automated_2017, hirose_spf-celltracker_2018, xu_spatial-temporal_2019, hossain_visual_2018}, and graph-based matching~\cite{padfield_coupled_2011, magnusson_global_2015, arbelle_probabilistic_2018, turetken_network_2017, schiegg_graphical_2015, akram_cell_2017, loffler_graph-based_2021}. The recent success of deep learning based cell segmentation, surpassing traditional segmentation approaches~\cite{stringer_cellpose_2021, cutler_omnipose_2021, scherr_cell_2020, weigert_star-convex_2020} and the successful application of deep learning for multi object tracking~\cite{pal_deep_2021, l_kalake_analysis_2021, ciaparrone_deep_2020}, give reason to apply deep learning to the task of cell tracking as well.
 
Currently, the number of approaches incorporating deep learning for cell tracking is still small. In~\cite{payer_segmenting_2019} and~\cite{zhao_voxelembed_2021} a Convolutional Neural Network (CNN) shaped like a stacked hourglass with convolutional gated recurrent units is used to predict pixel-wise embeddings, which are subsequently clustered into instances. Hayashida et al. train a single model to detect and track cells by predicting offsets and a magnitude which quantifies how likely a cell is at a specific position~\cite{hayashida_mpm_2020}. He et al. combine a particle filter with multi task learning to learn an observation model to select the most probable candidate in the next frame~\cite{he_cell_2017}. Hayashida and Bise train two separate models for detection and tracking and predict a motion flow of detected centroids between successive frame pairs~\cite{shen_cell_2019}. Wen et al. train a multi layer perceptron to predict similarity scores of pairs of already segmented cells between successive frames based on position and their Euclidean distances to their neighbors~\cite{wen_3deecelltracker_2021}. In~\cite{xie_deep_2021}, after applying a cell segmentation, pairs of image patches are forwarded to a CNN to predict similarity scores which are used to match cells using the Hungarian method. Chen et al. extend a Mask R-CNN with a Siamese branch for tracking, using handcrafted and learned features as embeddings to link corresponding cells~\cite{chen_celltrack_2021}. Sugawara et al. learn, based on sparse annotations, two separate models for cell detection and tracking, where for tracking flows between successive frames are estimated~\cite{sugawara_tracking_2021}. In~\cite{lugagne_delta_2020}, two separate models are trained for segmentation and tracking of cells, whereas the tracking model receives a pair of successive frames, a binary mask and the mask of the cell to track and learns to predict the mask of the cell in the next frame. Ben-Haim et al. use a cascade of approaches to track already segmented cells: first deep metric learning is used extract cell features, then a graph neural network learns to predict linking costs based on the extracted embeddings, finally, the cells are matched based on the linking costs using a graph-based matching~\cite{ben-haim_graph_2022}. To handle the lack of fully annotated data,~\cite{liu_towards_2021} trains a CycleGAN to generate synthetically annotated training data, whereas~\cite{nishimura_weakly-supervised_2020} learns to generate pseudo tracking labels from detection annotations.

Methods based on a cascade of models~\cite{he_cell_2017,shen_cell_2019, wen_3deecelltracker_2021, xie_deep_2021, sugawara_tracking_2021, lugagne_delta_2020, wang_deep_2020, ben-haim_graph_2022} can be complicated and can usually not be trained end to end, whereas approaches using only a single model~\cite{payer_segmenting_2019, zhao_voxelembed_2021, hayashida_mpm_2020, chen_celltrack_2021} can be trained at once. Approaches using recurrent network elements~\cite{payer_segmenting_2019, zhao_voxelembed_2021, he_cell_2017}, can need careful re-initialization or updates in contrast to models without recurrent elements~\cite{hayashida_mpm_2020,shen_cell_2019}. Embeddings can be estimated for all cells simultaneously, followed by calculating their pairwise distances afterwards~\cite{payer_segmenting_2019, zhao_voxelembed_2021, he_cell_2017, hayashida_mpm_2020, shen_cell_2019, sugawara_tracking_2021, nishimura_weakly-supervised_2020} or a model can predict similarity scores for distinct pairs of cells~\cite{wen_3deecelltracker_2021, chen_celltrack_2021, xie_deep_2021, lugagne_delta_2020, he_cell_2017}. Often, the learned embeddings are not human comprehensible~\cite{payer_segmenting_2019, zhao_voxelembed_2021, chen_celltrack_2021}, in contrast, learning embeddings that represent offsets are easy to interpret~\cite{hayashida_mpm_2020, shen_cell_2019, sugawara_tracking_2021, nishimura_weakly-supervised_2020}.

In summary, we can categorize recent deep learning based cell segmentation and tracking approaches by: (i) the complexity of their setup, e.g. needing several models for segmentation and tracking vs. a single model, (ii) their network architecture, e.g. if recurrent network elements are used, (iii) their learned output, e.g. predicting similarity scores or embeddings, and (iv) the type of the learned embedding, e.g. whether the embedding is comprehensible for humans.

Here, we propose a single CNN for cell segmentation and tracking, without any recurrent network elements learning human comprehensible embeddings. Our work is based on the instance segmentation approach of Neven et al.~\cite{neven_instance_2019}, which trains a branched ERFNet~\cite{romera_erfnet_2018} to learn the offset of pixels to its object center as well as a clustering bandwidth to cluster pixels into instance masks. Lalit et al. applied this approach first to the task of cell segmentation~\cite{lalit_embedding-based_2021}. We extend the instance segmentation approach to simultaneous segmentation and tracking, by including an estimation of offsets for cell pixels to their cell center between successive frames. Therefore, we adapt the branched ERFNet to process pairs of images and extend the network with an additional decoder branch for tracking. To show the competitiveness of our approach named EmbedTrack, we evaluate our approach on diverse 2D data sets of the well-established Cell Tracking Benchmark (CTB) from the Cell Tracking Challenge (CTC)~\cite{maska_benchmark_2014, ulman_objective_2017}. We provide our code at \href{https://git.scc.kit.edu/kit-loe-ge/embedtrack}{https://git.scc.kit.edu/kit-loe-ge/embedtrack}.

Our work has most similarity with the cell detection and tracking approach of~\cite{hayashida_mpm_2020}, in the sense that we train a CNN learning offsets. However, the approaches differ in the used network architecture, the loss function, and the post-processing, our approach is learning a clustering bandwidth and~\cite{hayashida_mpm_2020} learns a 3D embedding including position and time. Furthermore,~\cite{hayashida_mpm_2020} only detects cells as points while our network learns cell segmentation.

In the following, we start by introducing the instance segmentation approach of Neven et al.~\cite{neven_instance_2019}. Next, we show how to extend the instance segmentation approach to joint cell segmentation and tracking. In the experiments section, we participate as team KIT-Loe-GE on the CTB. Besides reporting our results on the benchmark, we provide insights into the cell statistics of the data sets concerning cell counts and cell motility as well as provide run time estimates of our algorithm.

\begin{figure}
    \centering
    \includegraphics[width=\linewidth]{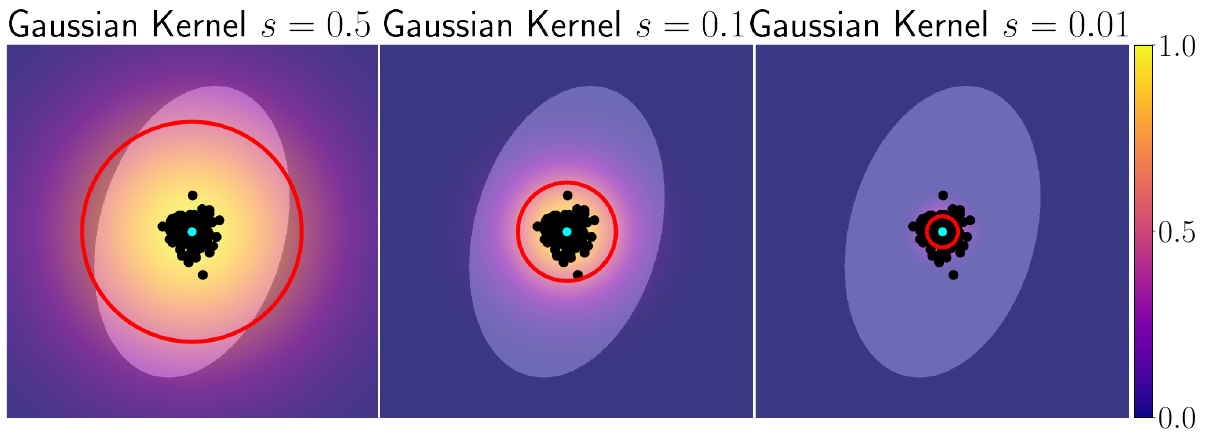}
    \caption{\textbf{Visualization of the clustering using different clustering bandwidths.} The plots show the distance $d(\vec{c}, \vec{p})$ using Gaussian kernels with different bandwidths $s=s_x=s_y$ in range $[0, 1]$ (heat maps from blue to yellow) of each pixel $\vec{p}$ in the plot to the object center $\vec{c}$ (cyan) of the ellipsoid object. The red circles are contours where the distance score between the center $\vec{c}$ and any pixel $\vec{p}$ is $d(\vec{c}, \vec{p})=0.5$. The points (black) visualize pixels shifted by their predicted offset to the center of the ellipsoid object. To cluster the pixels into instances, points (black) will be assigned to the same instance if they lay within the red circle.}
    \label{fig:kernel}
\end{figure}

\section{Instance Segmentation of Neven et al.}
Instance segmentation allows to distinguish between different instances of the same semantic class in an image. Therefore, foreground pixels need to be assigned to different instances, for instance by finding clusters of foreground pixels and assigning pixels belonging to the same cluster to the same instance mask. To cluster pixels, a clustering bandwidth, which is often defined manually, is needed. The clustering bandwidth determines in combination with a distance measure how far pixels can be apart from a cluster center to be still assigned to this cluster center. Neven et al. proposed the idea of training a CNN to predict offsets to shift pixels along their $x$- and $y$-dimension to their object centers, so they form compact clusters, as well as learning the clustering bandwidths $s$ required for each object, instead of selecting them manually~\cite{neven_instance_2019}. As distance measure, ~\cite{neven_instance_2019} use the Gaussian kernel
\begin{equation}
    d(\vec{c}, \vec{p}) = \exp\left(- \frac{\left(c_x - p_x\right)^2}{s_x} - \frac{\left(c_y - p_y\right)^2}{s_y}\right),
\end{equation}
where $\vec{c}$ and $\vec{p}$ are 2D position vectors, $c_x$ and $p_x$ the $x$-dimension of the vectors $c_y$ and $p_y$ the $y$-dimension of the vectors, and $s_x$ and $s_y$ are the bandwidth in $x$- and $y$-dimension. Using such a distance measure provides the possibility to allow for more slack in estimating the object center of large objects and avoid over-segmentation by increasing the bandwidth, whereas a more precise estimation of the object center of small objects can be enforced by reducing the bandwidth. Fig.~\ref{fig:kernel} shows distance measures based on the Gaussian kernel with different bandwidths $s$. A too large bandwidth leads to merging pixels belonging to different objects into one instance mask, which can result in merging of objects - under-segmentation error - whereas a too small bandwidth leads to not all shifted pixels being clustered to the same instance, which can result in splitting an object -- over-segmentation error.

As the true object center is unknown during inference, the idea is that by learning to predict the clustering bandwidths and offsets for each pixel, pixels can be clustered into instances by selecting any pixel, since its shift by the learned offset is an estimate of the object center and the clustering bandwidth defines how far other shifted pixels can be apart to be still assigned to the same instance.

\section{Method}
\begin{figure*}
    \centering
    \includegraphics[width=\linewidth]{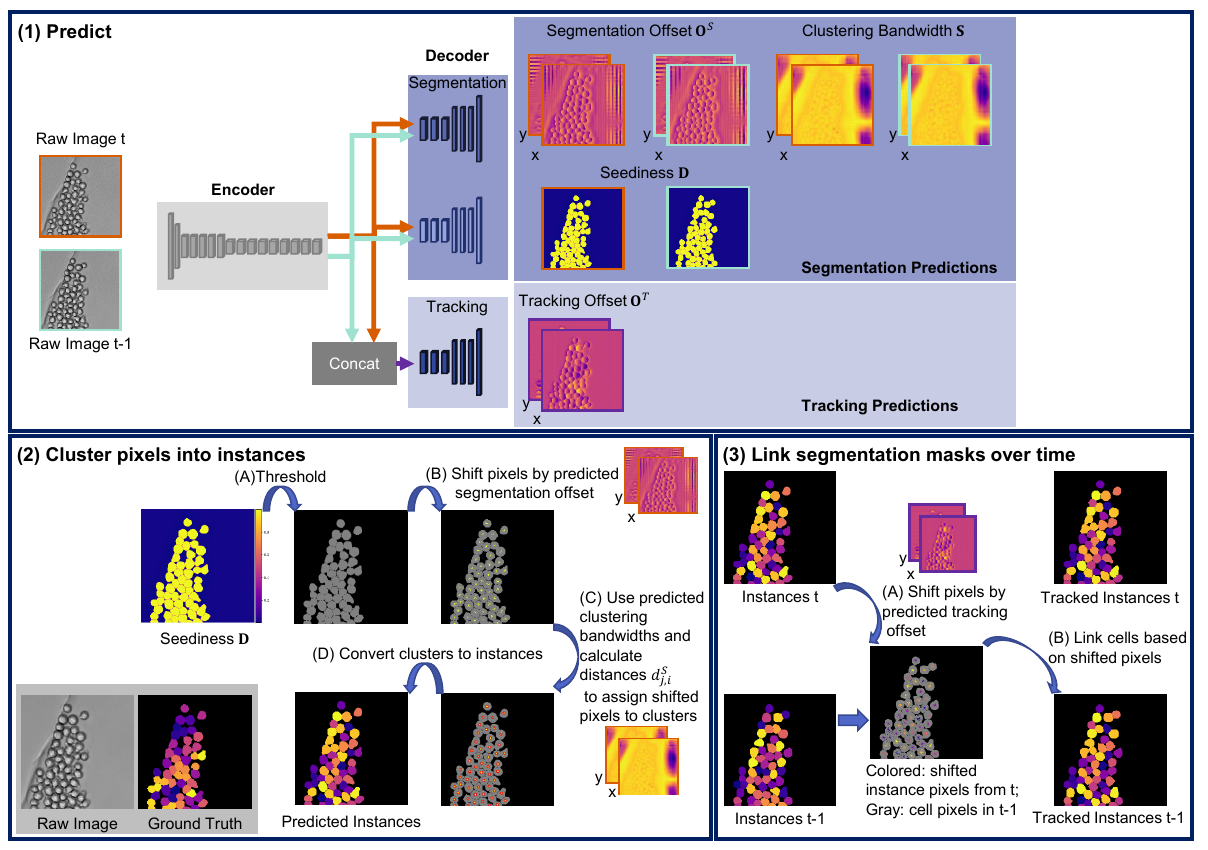}
    \caption{\textbf{Method Overview.} The approach consists of three steps: (1) prediction of offsets, clustering bandwidths and seediness maps using a CNN, (2) processing the predicted segmentation offsets and clustering bandwidths in a clustering step to retrieve an instance segmentation, and (3) linking the instance segmentation masks over time by processing the by the CNN predicted tracking offsets. (1) The CNN, a branched ERFNet, receives pairs of raw image crops, of time points $t$ and $t-1$. The CNN predicts two sets of segmentation predictions, one set for each of the two time points, and one tracking offset tensor. The offsets and the clustering bandwidths are tensors, where their values along the $x$-and $y$-dimension are shown as maps, where $H$ and $W$ are the height and the width of the raw image crops. (2) The segmentation predictions are processed in a clustering step to retrieve instance segmentation masks for time points $t$ and $t-1$, where gray and black show the prediction of the network for cell and background and the yellow pixels show the cell pixels after adding the predicted offset. The red circles indicate $d^s_{j,i}=0.5$ with the predicted clustering bandwidths. (3) The instance segmentation masks are linked backwards in time using the predicted tracking offsets, which is a tensor here shown with its values along $x$- and $y$-dimension. Raw image crops of the data set BF-C2DL-HSC from the CTC~\cite{maska_benchmark_2014, ulman_objective_2017}.}
    \label{fig:method_overview}
\end{figure*}

In the following, we show how to extend the instance segmentation approach of Neven et al. to cell segmentation and tracking by adding an additional decoder path to the CNN for tracking and introduce an additional tracking step. An overview of our approach is shown in Fig.~\ref{fig:method_overview}. First, pairs of image crops of the successive time points $t$ and $t-1$ are forwarded to a CNN, which predicts for each pixel segmentation offsets and clustering bandwidths for segmentation, and tracking offsets of pixels belonging to a cell at $t$ to their cell center at $t-1$ for tracking. Next, the segmentation offsets and clustering bandwidth predictions are processed with a foreground-background prediction in a clustering step, to assign pixels predicted as cell (foreground) to instances. Finally, based on the predicted tracking offsets between the two time points and the instance masks retrieved from the clustering step, the instances are linked backwards in time.

In the following, the network architecture, the loss function, the clustering step, and tracking step are explained in more detail.

\subsection{EmbedTrack Model}
We use a branched ERFNet~\cite{romera_erfnet_2018} with one shared encoder and three decoder paths -- two for segmentation and one for tracking. A visualization of the model and the predicted outputs is shown in Fig.~\ref{fig:method_overview}. An image pair is fed through the shared encoder, where the images are processed individually to omit entangling their feature representations. For tracking, the resulting feature representations are concatenated and forwarded to the tracking branch, whereas for segmentation the feature representations are forwarded to the segmentation branches separately. For each time point, $t$ and $t-1$, the network predicts a set of segmentation predictions, whereas for tracking one tracking offset prediction from $t$ to $t-1$ is predicted.

For raw images of shape $[H,W]$, where $H$ is the height and $W$ the width, the first segmentation decoder predicts segmentation offsets $\mat{O}^S$ of shape $[2, H, W]$ between pixels belonging to a cell and their corresponding cell center, and clustering bandwidths $\mat{S}$ of shape $[2, H, W]$. The second segmentation decoder predicts the seediness $\mat{D}$ of shape $[H, W]$, which is a score between $0$ and $1$ which serves as a foreground-background estimation as well as indicates if a cell pixel estimates its cell center correctly. The tracking decoder learns to predicts tracking offsets $\mat{O}^T$ of shape $[2, H, W]$ between pixels belonging to a cell at $t$ and their corresponding cell center at $t-1$. We apply a tanh activation on the segmentation offset and tracking offset prediction and a sigmoid activation on the clustering bandwidth prediction and the seediness prediction. In summary, the network predicts one set of segmentation predictions $\set{P}_{seg}=\{\mat{S}, \mat{O}^S, \mat{D}\}$ for each image $t$ and $t-1$, and one tensor of tracking offsets $\mat{O}^T$ to link pixels belonging to a cell at $t$ to its corresponding cell center at $t-1$.

\subsection{Loss}
The aim is to train the segmentation decoders of the CNN to predict a foreground-background prediction, segmentation offsets, and clustering bandwidths that are required for the subsequent clustering step, whereas the tracking decoder is trained to predict tracking offsets to link instance masks from $t$ to $t-1$. In the loss, the centers of the cell instances and the ground truth instance masks are used. We like to emphasize that the offsets and clustering bandwidths have no direct supervision during training.

Pixels belonging to the same instance should predict similar clustering bandwidths, since in the clustering step any shifted pixel could be selected as the cluster center and its clustering bandwidth defines the range in which other pixels will be assigned to the same cluster and therefore the same instance. To enforce similar clustering bandwidths~\cite{neven_instance_2019}, use a loss component based on the variance between the clustering bandwidth vectors belonging to the same instance. Therefore, for each instance $m$ the mean clustering bandwidth vector $\vec{\overline{s}}_m$, in $x$- and $y$-dimension, is calculated over the set of pixel indices $\set{P}_m$ belonging to the instance mask $m$, where $\vec{s}_{k}$ is the clustering bandwidth vector at pixel index $k$ of the clustering bandwidth prediction $\mat{S}$
\begin{equation}
    \vec{\overline{s}}_m = \frac{1}{|\set{P}_m|}\sum_{k \in \set{P}_m} \vec{s}_{k}.
\end{equation}
The loss part enforcing similar clustering bandwidth predictions for pixels belonging to the same masks is then given as
\begin{equation}
    \mathcal{L}_{var} = \frac{1}{M_{inst}} \sum_{m=1}^{M_{inst}} \frac{1}{|\set{P}_m|}\sum_{k \in\set{P}_m} (\vec{\overline{s}}_m - \vec{s}_{k})^2,
\end{equation}
where $M_{inst}$ is the number of instance masks and $|\set{P}_m|$ is the number of pixels of instance $m$.\\

To get the predictions of the cell centers for segmentation and tracking, the predicted offsets $\vec{o}^S_i$ for segmentation and $\vec{o}^T_i$ for tracking, where $i$ is a multi-index, referring to the pixel position in $x$- and $y$-dimension, are added to the position of each pixel
\begin{equation}
    \begin{aligned}
    \vec{e}^S_i &= \vec{p}_i +\vec{o}^S_i,\\
    \vec{e}^T_i &= \vec{p}_i +\vec{o}^T_i,\\
    \end{aligned}
\end{equation}
where $\vec{e}^S_i$ is the predicted cell center of pixel $i$ at the same time point, $\vec{e}^T_i$ is the predicted cell center of pixel $i$ from time point $t$ for its cell center at $t-1$, and $\vec{p}_i$ is the normalized pixel position -- pixel coordinates from range $([0, H], [0,W])$ normalized to range $([0,1], [0,1])$. By using the already introduced distance measure, we get for each pixel index $i$ a prediction how close its shifted position $\vec{e}^S_i$ is to the cell center
\begin{equation}
    \begin{aligned}
      \vec{s}_m &= \exp(w_{s}\cdot\vec{\overline{s}}_m),\\
   d^S_{m,i}\left(\vec{c}_{m},\vec{e}^S_i\right)  &= \exp\left(-\frac{\left(c_{mx}-e^S_{ix}\right)^2}{s_{mx}} - \frac{\left(c_{my}-e^S_{iy}\right)^2}{s_{my}}\right), \\ 
    \end{aligned}
\end{equation}
where $d^S_{m,i}(\vec{c}_{m},\vec{e}^S_i)$ is the distance between the cell center $\vec{c}_{m}$ of instance mask $m$ and the predicted cell center $\vec{e}^S_i$, $s_{mx}$ and $s_{my}$ are the $x$- and $y$-dimensions of a scaled mean clustering bandwidth vector $\vec{s}_m$ and $w_s$ a scaling weight which we set to $-10$. As cell center $\vec{c}_m$ for each instance mask $m$ we choose as in~\cite{lalit_embedding-based_2021} the medoid, so the cell center will always lay inside of the cell. 

In~\cite{neven_instance_2019} the segmentation offsets and clustering bandwidths are learned jointly using an instance loss based on the Lovász hinge loss~\cite{berman_lovasz-softmax_2018, yu_learning_2015}. The Lovász hinge loss is a convex surrogate for sub-modular losses which allows an efficient minimization of sub-modular loss functions such as the Jaccard loss~\cite{yu_learning_2015}. The Jaccard loss is given by 
\begin{equation}
    \mathcal{L}_{Jaccard} = 1 - J(\set{S}_{pred},\set{S}_{GT}),
\end{equation}
where $J(\set{S}_{pred},\set{S}_{GT})$ is the Jaccard index between the set of pixels $\set{S}_{pred}$ belonging to the predicted instance mask and the set of pixels $\set{S}_{GT}$ belonging to the ground truth instance mask. By minimizing the Lovász hinge loss, the predicted segmentation offsets and clustering bandwidths can be jointly optimized such that the Jaccard index between predicted instance mask and ground truth instance mask will be maximized
\begin{equation}
\begin{aligned}
\mat{B}_{m} &= \left(b_{m,i}\right) = \begin{cases}
 1 \quad \text{if} \; i \in \set{P}_{m}\\
 0 \quad \text{otherwise}\\
\end{cases},\\
    \mat{D}^{S}_{m} &=  (d^S_{m,i}(\vec{c}_m,\vec{e}^S_i)),\\
        \mathcal{L}_{instance} &= \sum_{m=1}^{M_{inst}}\mathcal{L}_{Lov\Acute{a}sz}\left(2\cdot\mat{D}^{S}_{m}-\mat{1}, 2\cdot\mat{B}_m-\mat{1}\right),
\end{aligned}
\end{equation}
where, $\mat{B}_m$ is a binary mask which is $1$ at indices belonging to the instance mask $m$ and $0$ otherwise, $\mat{D}^S_m$ is a distance matrix of shape $[H,W]$ containing the distance of each shifted pixel to the cell center $\vec{c}_m$ of instance mask $m$, $\mat{1}$ is a matrix of ones the same size as $\mat{D}^S_m$, and $\mathcal{L}_{Lov\Acute{a}sz}$ the Lovász hinge loss.

\begin{equation}
\begin{aligned}
\mat{B}_{m} &= \left(b_{m,i}\right) = \begin{cases}
 1 \quad \text{if} \; i \in \set{P}_{m}\\
 0 \quad \text{otherwise}\\
\end{cases},\\
    \mat{D}^{S}_{m} &=  (d^S_{m,i}(\vec{c}_m,\vec{e}^S_i)),\\
\end{aligned}
\end{equation}
where, $\mat{B}_m$ is a binary mask which is $1$ at indices belonging to the instance mask $m$ and $0$ otherwise, $\mat{D}^S_m$ is a distance matrix of shape $[H,W]$ containing the distance of each shifted pixel to the cell center $\vec{c}_m$ of instance mask $m$, and $\mat{1}$ is a matrix of ones the same size as $\mat{D}^S_m$.

To cluster pixels into instances, first the foreground pixels need to be selected, therefore, a foreground-background prediction is needed. The seediness map $\mat{D}$, which the second segmentation decoder is learning, predicts for each pixel if it belongs to background or a cell, and if the pixel belongs to a cell how close the predictions of the cell center $\vec{e}^S_i$ is to the actual cell center. It therefore serves as a foreground-background estimation and as a prediction of how well pixels estimate their cell center. To learn the seediness map, an additional seed loss part is used in~\cite{neven_instance_2019}. For the seediness map, the CNN learns to predict for each pixel $i$ the distance measure $d^S_{m,i}(\vec{c}_{m},\vec{e}^S_i)$ for pixels belonging to a cell and regressing to $0$ for background pixels
\begin{equation}
\begin{aligned}
   \mathcal{L}_{seed} &= w_{fg} \sum_{m=1}^{M_{inst}} \frac{1}{|\set{P}_m|} \sum_{k \in \set{P}_m} (d^S_{m,k}(\vec{c}_{m},\vec{e}^S_k) - d_k)^2 \\ 
   &+ \frac{1}{|\set{P}_{bg}|}\sum_{j \in \set{P}_{bg}} (d_j - 0)^2,
   \end{aligned}
\end{equation}
where $d_k$ is the predicted seediness score of pixel $k$ belonging to the set of pixel indices $\set{P}_m$ of instance mask $m$, $d_j$ is the predicted seediness score at background pixel $j$ in the seediness map, $\set{P}_{bg}$ the set of background pixel indices, and $w_{fg}$ is a weight for foreground pixels.

The segmentation loss is then given as
\begin{equation}
    \mathcal{L}_{seg} = w_{instance}\mathcal{L}_{instance} + w_{var}\mathcal{L}_{var} + w_{seed}\mathcal{L}_{seed},
\end{equation}
where the weights are set to $w_{instance}=w_{seed}=w_{fg}=1$, and $w_{var}=10$. Since the network provides segmentation predictions for pairs of images, the segmentation loss is calculated for both images of time points $t$ and $t-1$ separately and accumulated.

For tracking, the aim is to shift pixels belonging to a cell at $t$ to their cell center at $t-1$. Similar to the instance segmentation loss, we propose to use the Lovász hinge loss for tracking. Since the network predicts two sets of segmentation predictions, one for time point $t$ and one for time point $t-1$, we add time indices in the following to highlight to which time point the predicted segmentation components belong. $d^T_{m,i}(\vec{c}_m^{t-1},\vec{e}^T_i)$ is the distance between $\vec{c}_m^{t-1}$, the cell center of instance $m$ at time point $t-1$, and $\vec{e}^T_i$, which is the predicted cell center of pixel $i$ belonging to time point $t$ shifted to its cell center at $t-1$
\begin{equation}
    \begin{aligned}
   d^T_{m,i}\left(\vec{c}_m^{t-1},\vec{e}^T_i\right)  &= \exp\left(-\frac{\left(c^{t-1}_{mx}- e^T_{ix}\right)^2}{s^{t}_{mx}} - \frac{\left(c^{t-1}_{my}- e^T_{iy}\right)^2}{s^{t}_{my}}\right),\\
 \mat{D}^{T}_{m} &=  \left(d^T_{m,i}\left(\vec{c}^{t-1}_m,\vec{e}^T_i\right)\right),\\
 \mat{B}_{m}^t &= \left(b_{m,i}^t\right) = \begin{cases}
 1 \quad \text{if} \; i \in \set{P}^{t}_{m}\\
 0 \quad \text{otherwise}\\
 \end{cases},\\
 \mathcal{L}_{track} &= \sum_{m=1}^{M^t_{inst}}\mathcal{L}_{Lov\Acute{a}sz}\left(2\cdot\mat{D}^{T}_{m}-\mat{1}, 2\cdot\mat{B}^t_m-\mat{1}\right),
    \end{aligned}
\end{equation}
where $\set{P}^t_m$ are the pixel indices of the instance segmentation mask $m$ at time point $t$, $\mat{D}^T_m$ is a distance matrix of shape $[H,W]$ containing the distance of each shifted pixel to the cell center $\vec{c}^{t-1}_m$ of instance mask $m$ at time point $t-1$, $\mat{1}$ is a matrix of ones the same size as $\mat{D}^T_m$, and $\mat{B}^t_m$ is a binary mask of instance $m$ at time point $t$, and $M^t_{inst}$ is the number of instances at time point $t$.

The loss is finally given as the sum of segmentation and tracking loss
\begin{equation}
    \mathcal{L} = w_{seg}\mathcal{L}_{seg} + w_{track}\mathcal{L}_{track},
\end{equation}
where $w_{seg}$ and $w_{track}$ are weights both set to $1$.

\begin{figure*}
\includegraphics[width=\linewidth]{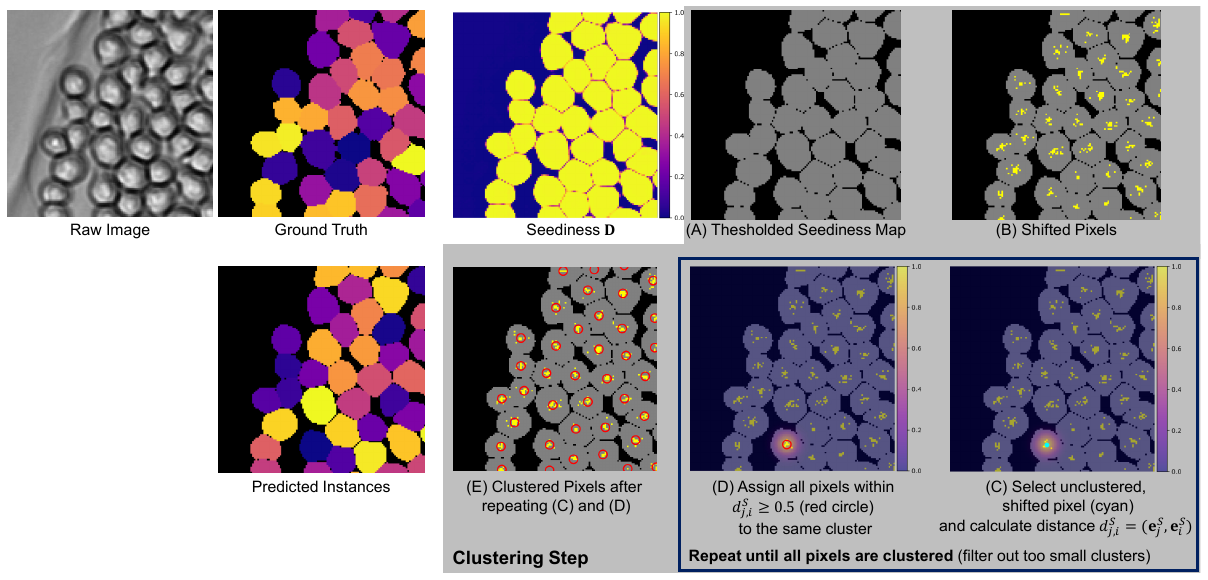}
    \caption{\textbf{Clustering step}. (A) the seediness map $\mat{D}$ is thresholded resulting in a foreground (gray) background (black) prediction. Next, the pixels predicted as foreground are shifted by the predicted segmentation offsets $\mat{O}^S$, where the shifted pixels are shown in yellow (B). (C) The shifted pixels are assigned to clusters by selecting an unclustered, shifted pixel (cyan) and calculating its distance $d^S_{j,i}(\vec{e}^S_j, \vec{e}^S_i)$ to all other pixels $i$-- distance map shown as a heat map. (D) The distance map is thresholded and pixels with a distance score higher than $0.5$ (red circle) are assigned to the same cluster. The steps (C) and (D) are repeated until all pixels are clustered (E). Finally, the clustered pixels are converted into a instance masks, where each cluster represents an instance. Raw image crop of the data set BF-C2DL-HSC from the CTC~\cite{maska_benchmark_2014, ulman_objective_2017}.}
    \label{fig:clustering}
\end{figure*}

\subsection{Pixel Clustering}
To convert the pixel-wise predictions from the segmentation decoders to an instance segmentation, a clustering step is applied. We modify the clustering proposed in~\cite{neven_instance_2019} as follows: First, the clustering bandwidth tensor is smoothed with a 3x3 kernel along $x$- and $y$- dimension and will be referred to as $\mat{S}^{smooth}$. Since the seediness map $\mat{D}$ estimates how well pixels predict their cell center, we select pixels with scores larger than $0.5$ in the seediness map and compute the shifted pixel positions $\vec{e}_i^S$ to find potential cell centers. We convert each $\vec{e}_i^S$, which is a normalized pixel position in range $([0,1],[0,1])$ to an pixel index in range $([0,H], [0,W])$, accumulate them, and find potential cell centers by selecting pixel indices with more than five clustered pixels in their 3x3 neighborhood. Next, the pixel indices $j$ that refer to potential cell centers are sorted by their seediness map score $\vec{d}_j$. Then, starting with the pixel with the highest score, the list of likely cell centers is processed and pixels $i$ are assigned to the cell centers $\vec{e}^S_j$ using a similar distance measure as in the loss
\begin{equation}
\begin{aligned}
   \vec{s}_j &= \exp(w_{s}\cdot\mat{S}^{smooth}[j]),\\
    d^S_{j,i}(\vec{e}^S_j,\vec{e}^S_i)  &= \exp\left(-\frac{\left(e^S_{jx}- e^S_{ix}\right)^2}{s_{jx}} - \frac{\left(e^S_{jy}- e^S_{iy}\right)^2}{s_{jy}}\right),\\
    \mat{B}_j &= \begin{cases}
 1 \quad \text{if} \; \vec{d}^S_{j,i} > 0.5\\
 0 \quad \text{otherwise}
 \end{cases},
\end{aligned}
\end{equation}
where $\vec{s}_j$ is as in the loss a scaled clustering bandwidth vector at pixel $j$ with the same weight $w_{s}$ as used in the loss, $s_{jx}$ and $s_{jy}$ the x and y components of the clustering bandwidth vector, and $\mat{B}_j$ the resulting instance mask. Pixel indices referring to potential cell centers that are assigned to an instance mask will be removed from the list of likely cell centers. To filter out false positives, we set the minimum size of an instance mask to half the size of the 1\% percentile of all mask sizes of the training data set. On cell boundaries, pixels can have high distance scores for several cell centers. To assign the pixel to the best cell center, we keep for each pixel its highest distance score $d^S_{j,i}$ to allow to reassign a pixel to a subsequent cell center. Therefore, three conditions need to be full filled: the pixel receives a higher score $d^S_{j,i}$ if it is assigned to the new cell center, the number of pixels clustered into the new instance have at least the minimum size, and the fraction of pixels that are already assigned to another mask is less than half of the final instance mask. A visualization of the clustering step is shown in Fig.~\ref{fig:clustering}.

\subsection{Tracking}
After the clustering step, the instance segmentation masks are linked over time. Therefore, the shifted positions $\vec{e}^T_i$ are calculated for all pixels $i$ that belong to an instance mask at time point $t$ to find their corresponding cell centers at $t-1$. Each instance mask at $t$ is marked as a potential matching candidate for the instance mask at $t-1$ that contains the most shifted pixels of it. If a mask at $t-1$ has exactly one matching candidate, the two instance masks at $t$ and $t-1$ are assigned to the same track, whereas if an instance mask at $t-1$ has two potential matching candidates at $t$, the instance masks at $t$ are set as successors of the instance mask at $t-1$. In all other cases, shifted pixels not overlapping with any mask, more than two matching candidates - the mask at $t$ is marked as a new track starting at $t$.

\section{Experiments}
All models where trained on a system with Ubuntu18.04, an Intel i9 99000k, 32GB RAM and two Titan RTX with 24GB VRAM each. The approach was implemented in Python with PyTorch as deep learning framework.

\subsection{Data Sets}
\begin{figure*}
\begin{subfigure}[b]{0.19\linewidth}
\centering
\includegraphics[height=.12\textheight]{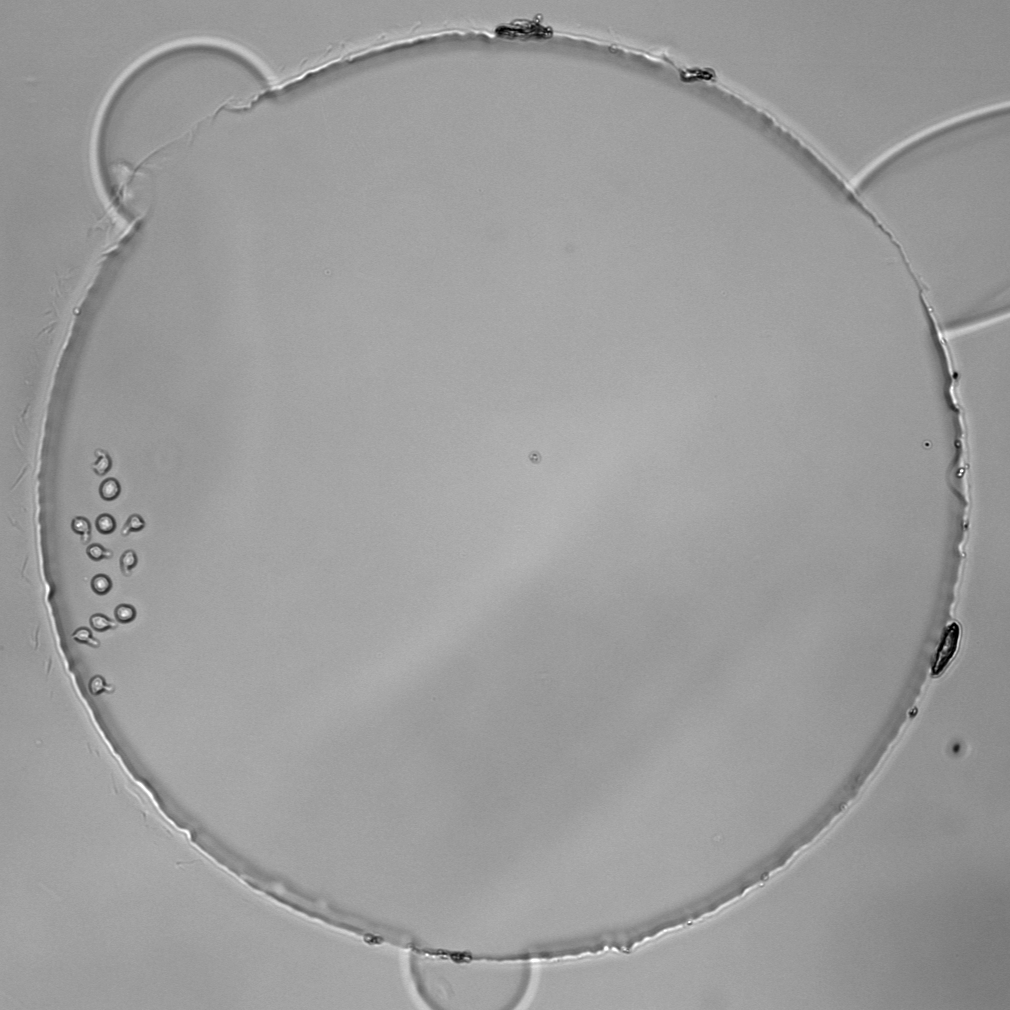}
\caption{BF-C2DL-HSC}
\label{fig:BF-C2DL-HSC}
\end{subfigure}%
\hfill
\begin{subfigure}[b]{0.19\linewidth}
\centering
\includegraphics[height=.12\textheight]{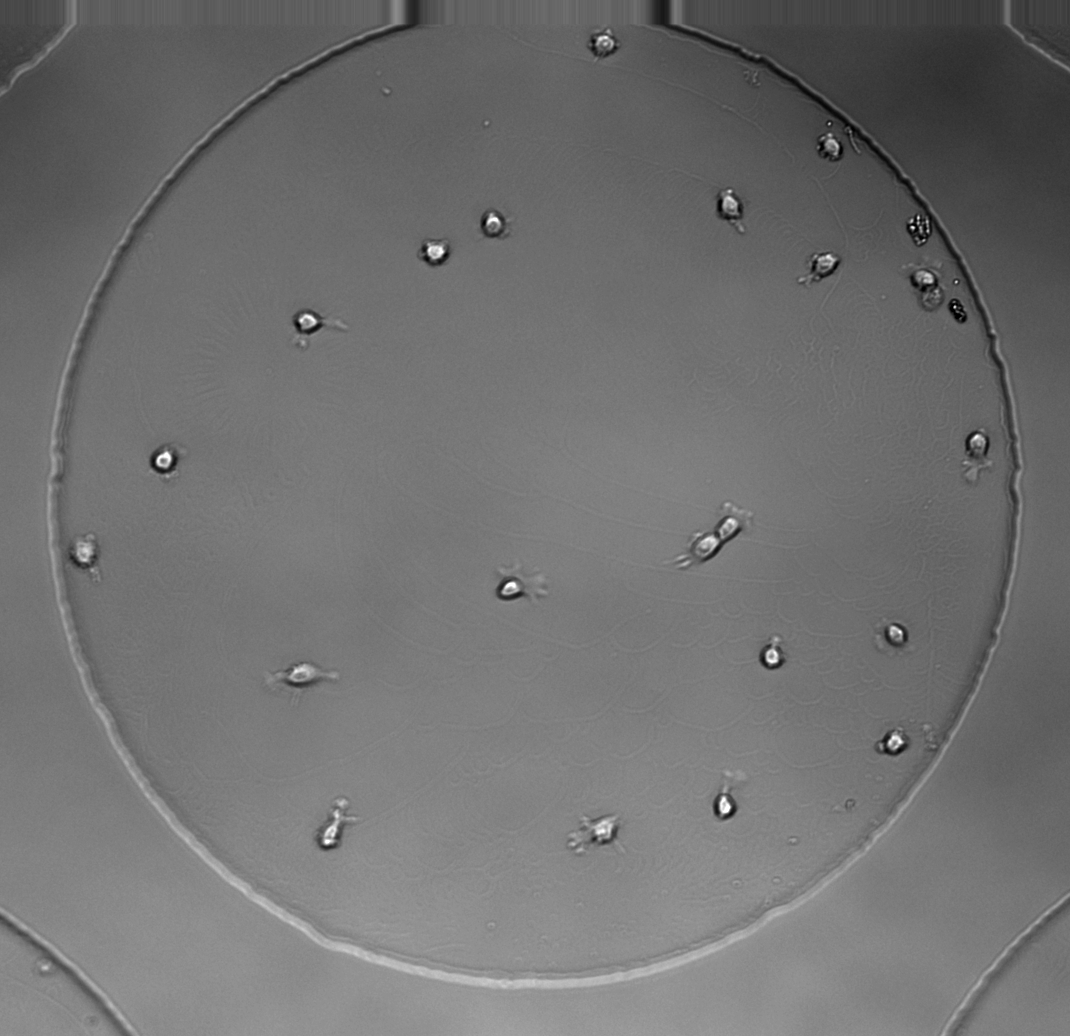}
\caption{BF-C2DL-MuSC}
\label{fig:BF-C2DL-MuSC}
\end{subfigure}%
\hfill
\begin{subfigure}[b]{0.19\linewidth}
\centering
\includegraphics[height=.12\textheight]{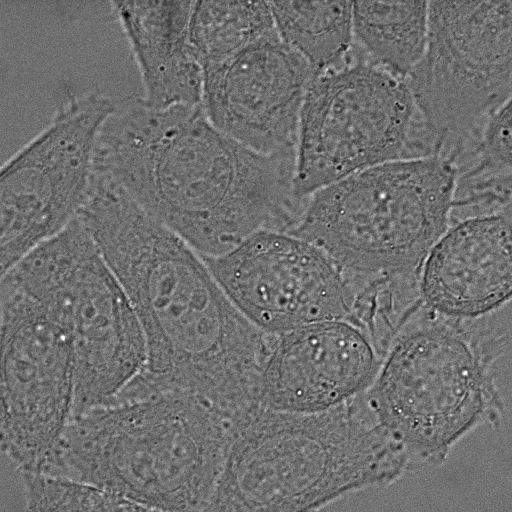}
\caption{DIC-C2DH-HeLa}
\label{fig:DIC-C2DH-HeLa}
\end{subfigure}%
\hfill
\begin{subfigure}[b]{0.19\linewidth}
\centering
\includegraphics[height=.12\textheight]{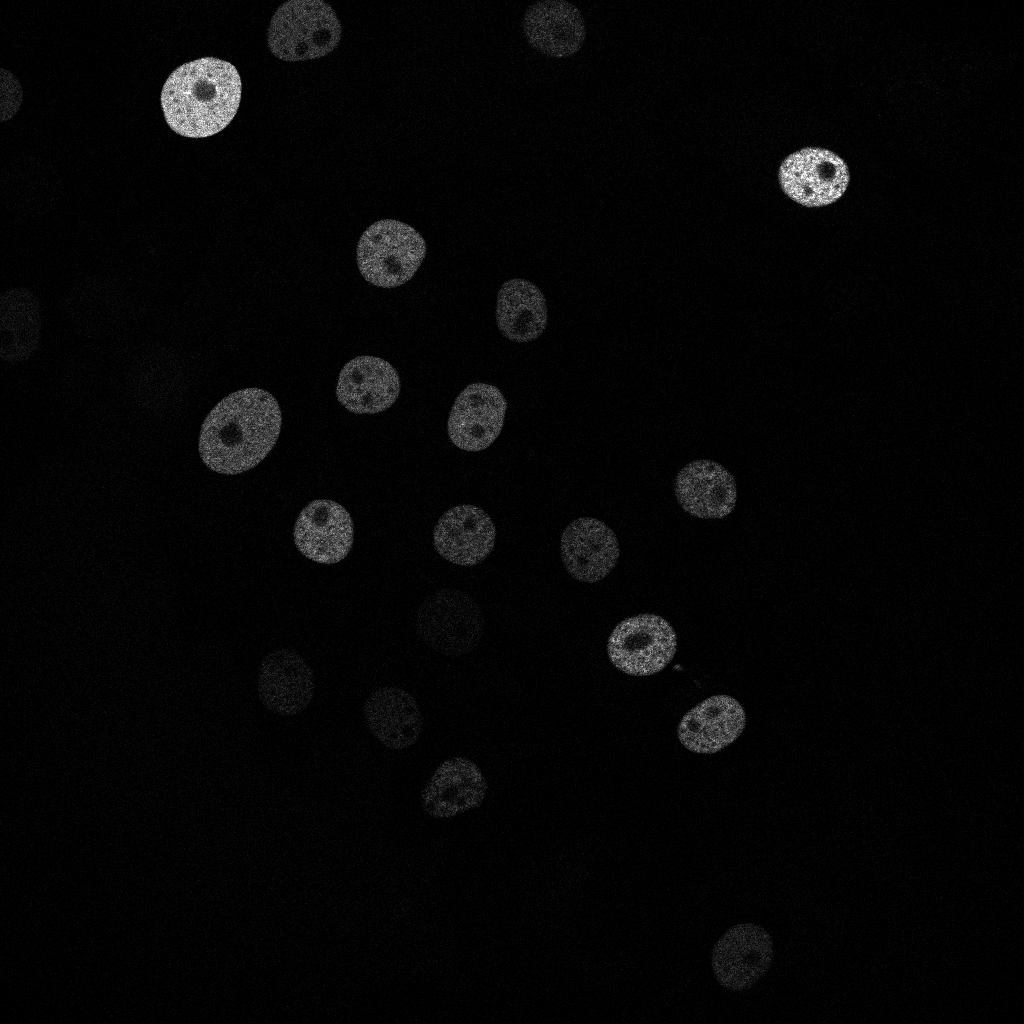}
\caption{Fluo-N2DH-GOWT1}
\label{fig:Fluo-N2DH-GOWT1}
\end{subfigure}
\hfill
\begin{subfigure}[b]{0.19\linewidth}
\centering
\includegraphics[height=.12\textheight]{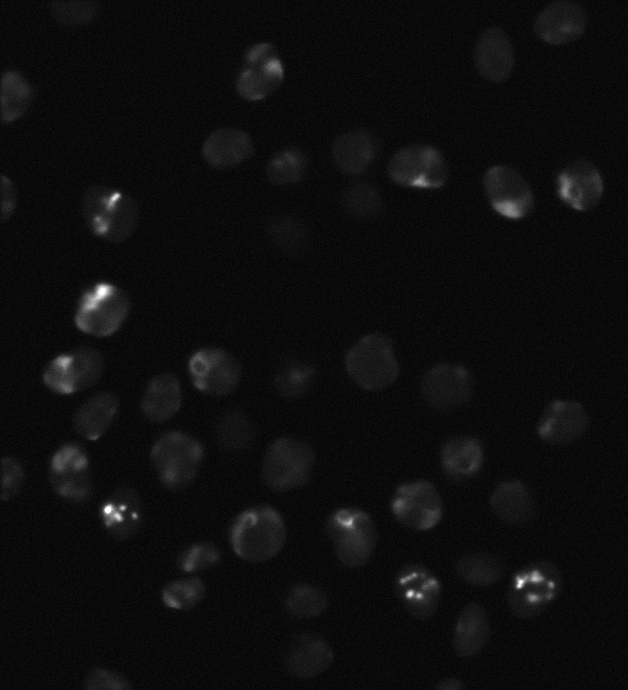}
\caption{Fluo-N2DH-SIM+}
\label{fig:Fluo-N2DH-SIM+}
\end{subfigure}\\
\begin{subfigure}[b]{0.19\linewidth}
\centering
\includegraphics[height=.12\textheight]{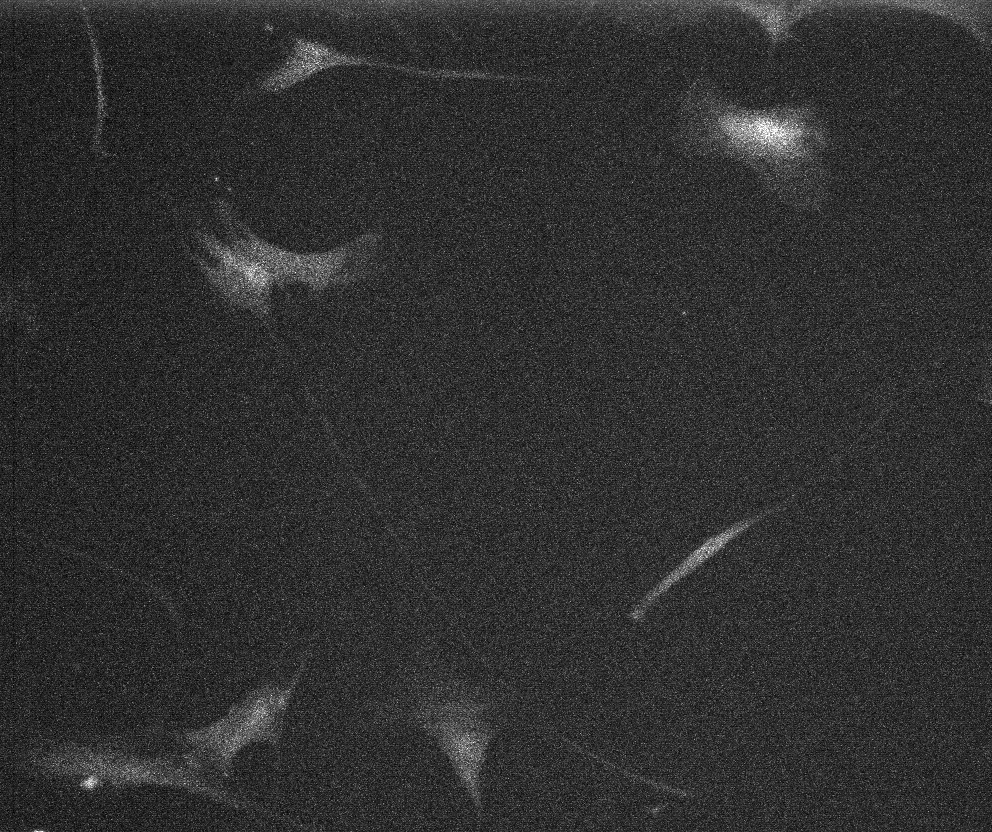}
\caption{Fluo-C2DL-MSC}
\label{fig:Fluo-C2DL-MSC}
\end{subfigure}%
\hfill
\begin{subfigure}[b]{0.19\linewidth}
\centering
\includegraphics[height=.12\textheight]{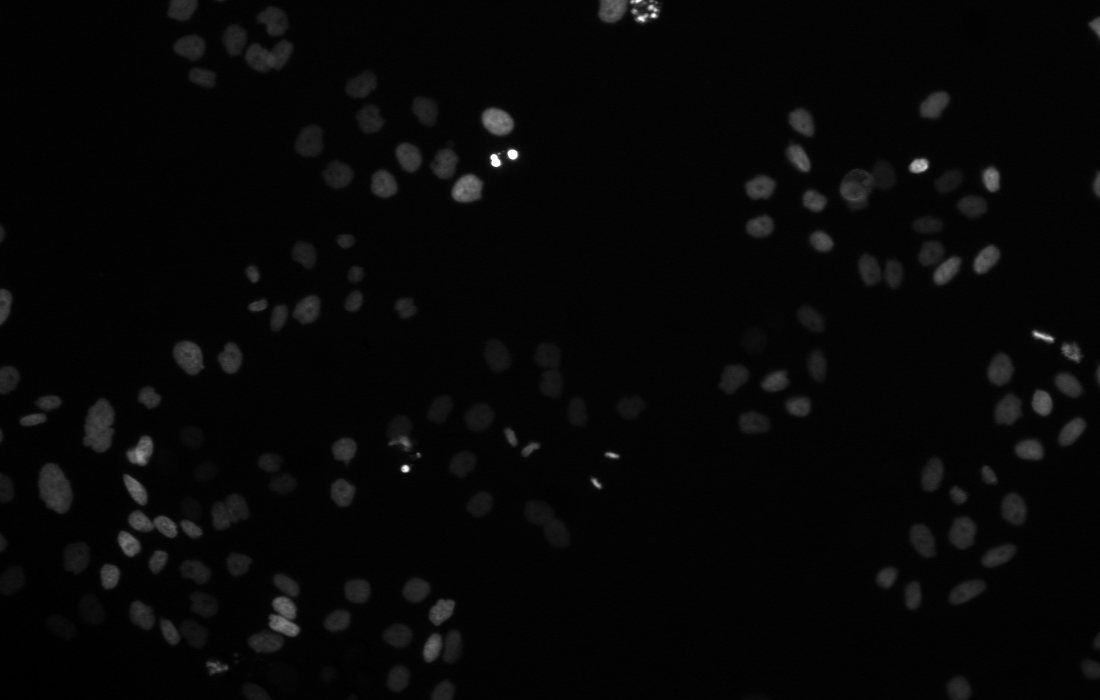}
\caption{Fluo-N2DL-HeLa}
\label{fig:Fluo-N2DL-HeLa}
\end{subfigure}%
\hfill
\begin{subfigure}[b]{0.19\linewidth}
\centering
\includegraphics[height=.12\textheight]{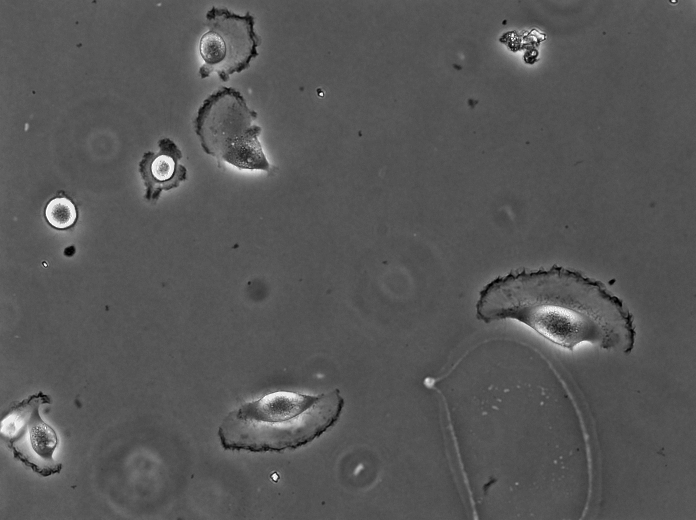}
\caption{PhC-C2DH-U373}
\label{fig:PhC-C2DH-U373}
\end{subfigure}%
\hfill
\begin{subfigure}[b]{0.19\linewidth}
\centering
\includegraphics[height=.12\textheight]{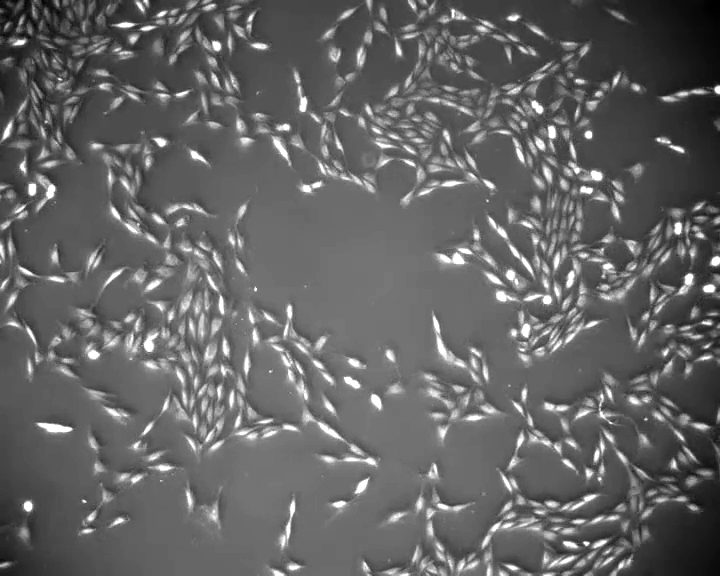}
\caption{PhC-C2DL-PSC}
\label{fig:PhC-C2DL-PSC}
\end{subfigure}%
    \caption{\textbf{Benchmark Data from the CTC.} All selected data sets from the CTC on which the EmbedTrack approach was evaluated and trained on. The contrast of the raw images has been adapted using min-max scaling to the percentiles 1 and 99 for visualization purposes. All data sets can be retrieved from \protect\url{http://celltrackingchallenge.net/}.}
    \label{fig:datasets}
\end{figure*}

The CTC provides two benchmarks based on the same cell data sets: the Cell Segmentation Benchmark (CSB) for segmentation and the Cell Tracking Benchmark (CTB) for tracking. We train EmbedTrack on nine 2D data sets from the CTC, which are shown in Fig.~\ref{fig:datasets}. The CTC data consists of two labeled images sequences for training, which we will refer to as training sequences 01 and 02, and two unlabeled image sequences for testing, which we will refer to as challenge sequences 01 and 02. For the training sequences, manually generated, Gold Truth (GT), annotations are provided for some segmentation masks, whereas the tracking information is fully provided as point-wise annotations for the nine selected data sets. All of the selected data sets, apart from Fluo-N2DH-SIM+ which is simulated and hence has a fully annotated GT, also contain a Silver Truth (ST). The ST are segmentation masks obtained from averaging predictions of previous submissions to the CTC and are provided by the CTC.

\subsection{Evaluation Metrics}
To compare the submitted cell segmentation and tracking algorithms, the CTC uses the metrics SEG, DET and TRA, which all lay in range $[0,1]$, where a higher score corresponds with a better performance. The SEG metric is based on the Jaccard similarity index, which calculates the ratio between the intersection of ground truth mask and predicted mask and the union of ground truth mask and predicted mask. Ground truth masks are matched to predicted masks, if the intersection between ground truth mask and predicted mask covers at least half of the ground truth mask. Ground truth masks without a matched predicted mask are penalized with a score of $0$. The DET and TRA score are based on the acyclic oriented graph matching (AOGM) measure~\cite{matula_cell_2015}. The AOGM penalizes the number of transformations needed to transform the predicted tracking graph into the ground truth tracking graph. The measure penalizes errors concerning the detection: false positives, false negatives, and merged cells and tracking: missing links, wrong links and links with wrong semantics. The DET score only includes errors concerning detection, whereas TRA includes all penalties for detection and tracking. On the CTB, the tracking algorithms are compared on the metrics SEG, TRA, and OP\textsubscript{CTB} which is the average between SEG and TRA score.

\subsection{Training}
We merge the segmentation masks from the ST with the point-wise annotations of the tracking GT to create fully labeled training data with reasonable annotation accuracy. To train our approach, we split the training sequences, 01 and 02, keeping the first 90\% of each image sequence for training and the last 10\% of each image sequence for evaluation. For submission to the CTB, we select the model with the best Intersection over Union (IoU) score on the evaluation data set. During training, we generate overlapping crops of size $256$x$256$ ($512$x$512$ for Fluo-C2DL-MSC). As augmentations during training we use CLAHE, blur, rotation, flipping and add shifts to simulate larger cell movement. Each image crop is normalized to range $[0, 1]$ using the $1\%$ and $99\%$ percentiles per image. We train each model for 15 epochs using the Adam optimizer~\cite{kingma_adam_2017} with learning rate $5\cdot10^{-4}$ and a one cycle learning rate scheduler.

\subsection{Inference}
For inference, we generate overlapping crops of of size $256$x$256$ ($512$x$512$ for Fluo-C2DL-MSC). We apply as test time augmentation rotation and flipping and min-max normalize each crop to range $[0, 1]$ using the $1\%$ and $99\%$ percentiles per image, then we forward the augmented crops to the trained model. We calculate the average over the augmented outputs and stitch the averaged predictions to predictions covering the entire image. Next, the clustering step generates the instance segmentation on the segmentation predictions of the entire image. Finally, we link instances to tracks by processing the tracking offsets and the instance segmentation masks as explained in the tracking step.

\begin{table*}
    \centering
      \caption{\textbf{Results of the Cell Tracking Benchmark (CTB)} -- status 14.03.2021. We participated as team KIT-Loe-GE on \protect\url{http://celltrackingchallenge.net/latest-ctb-results/}. Highlighted in bold are top 3 rankings, whereas bold and underlined indicates top 1 rankings. $T_{\text{inference}}$ per sequence is the average time needed to process the challenge data sets from loading the raw images from disk to saving the predicted tracking masks, averaged both sequences. By dividing the average run time through the number of frames per sequence, we get a rough estimate for processing a single frame. The cell statistics, cell counts and the overlap of the $10\%$ most motile cells calculated from the ST annotations of the training data sets.)}
    \label{tab:ctc_results}
    \resizebox{\textwidth}{!}{
   \begin{tabular}{c |c c c c c| c c c c}
    \hline
    &\multicolumn{5}{c}{\textbf{Challenge Data Results}}&\multicolumn{4}{c}{\textbf{Training Data Statistics}}\\
    \textbf{Data Set} & \textbf{SEG}&\textbf{TRA}&\textbf{OP\textsubscript{CTB}} & \multicolumn{2}{c}{\textbf{T\textsubscript{inference}}}& \multicolumn{2}{c}{\textbf{N Cells Last Frame}} & \multicolumn{2}{c}{\textbf{Overlap fastest 10\% Cells}}\\
    &(SEG Rank)&(TRA Rank)&(OP\textsubscript{CTB} Rank)&per sequence&per frame&01&02&01&02\\
    \hline
    \underline{\textbf{BF-C2DL-HSC}}&\makecell{\underline{\textbf{0.826}} \\\underline{\textbf{(1/12)}}}&\makecell{\underline{\textbf{0.985}}\\\underline{\textbf{(1/12)}}}&\makecell{\underline{\textbf{0.906}}\\\underline{\textbf{(1/12)}}}&$\sim$55:30min&1.9s&12&159&0.388&0.553\\
    \underline{\textbf{BF-C2DL-MuSC}}&\makecell{\underline{\textbf{0.782}}\\\underline{\textbf{(1/12)}}}&\makecell{\underline{\textbf{0.974}}\\\underline{\textbf{(1/12)}}}&\makecell{\underline{\textbf{0.878}}\\\underline{\textbf{(1/12)}}}&$\sim$43:00min&1.9s&23&22&0.098&0.275\\
    DIC-C2DH-HeLa&\makecell{0.824 \\ (6/22)}&\makecell{0.934 \\ (6/22)}&\makecell{0.879\\(6/22)}&$\sim$2:30min&1.8s&18&17&0.774&0.848\\
    Fluo-C2DL-MSC&\makecell{0.579\\(9/28)}&\makecell{0.693\\(11/28)}&\makecell{0.636\\(9/28)}&$\sim$2:20min&2.9s&8&3&0.709&0.031\\
    \textbf{Fluo-N2DH-GOWT1}&\makecell{\textbf{0.929}\\\textbf{(2/37)}}&\makecell{0.951\\(4/37)}&\makecell{\textbf{0.940}\\\textbf{(2/37)}}&$\sim$3:00min&2s&20&28&0.927&0.909\\
    \textbf{Fluo-N2DL-HeLa}&\makecell{0.906\\(4/36)}&\makecell{\textbf{0.992}\\\textbf{(2/36)}}&\makecell{0.949\\(4/36)}&$\sim$2:50min&1.9s&137&363&0.721&0.759\\
    \textbf{PhC-C2DH-U373}&\makecell{0.920\\(8/27)}&\makecell{\textbf{0.982}\\\textbf{(2/27)}}&\makecell{\textbf{0.951}\\\textbf{(2/27)}}&$\sim$2:00min&1.1s&7&5&0.875&0.881\\
    \textbf{PhC-C2DL-PSC}&\makecell{\textbf{0.740}\\\textbf{(2/28)}}&\makecell{\underline{\textbf{0.968}}\\ \underline{\textbf{(1/28)}}}&\makecell{\textbf{0.854}\\ \textbf{(2/28)}}&$\sim$7:35min&1.5s&661&498&0.862&0.875\\
    \underline{\textbf{Fluo-N2DH-SIM+}}&\makecell{\underline{\textbf{0.830}}\\\underline{\textbf{(1/36)}}}&\makecell{\underline{\textbf{0.979}}\\ \underline{\textbf{(1/36)}}}&\makecell{\underline{\textbf{0.905}}\\\underline{\textbf{(1/36)}}}&$\sim$2:40min&1.7s&45&54&0.863&0.841\\
    \hline
    \end{tabular}}
\end{table*}

\subsection{Evaluation on the CTB}
To evaluate the performance of our approach on diverse 2D data sets, we train and submit a model for each of the nine selected 2D data sets from the CTC. The results are shown in Table~\ref{tab:ctc_results}, where the metric scores SEG, TRA and OP\textsubscript{CTB} are the averages over the two challenge sequences. In addition, we report the time needed to process the two challenge sequences on our system, averaged over both challenge sequences, which includes all processing steps from loading the model, generating image crops, inferring them with the trained model, clustering, tracking and saving the predicted masks to disk. To provide additional insight on the data sets, we calculate the cell counts and cell motility based on the training sequences $01$ and $02$. As measure for cell motility, we use the ST annotations and calculate the overlap of cells between successive frames as their intersection of the two masks divided by the smallest size of the two masks and report the overlap for the $10\%$ of the most motile cells. For instance, the most motile $10\%$ of
the cells in the $01$ training sequence of BF-C2DL-HSC have an overlap smaller than 0.388. Based on the cell statistics, we identify challenging data sets based on their cell motility and their cell count. BF-C2DL-HSC and BF-C2DL-MuSC are data sets that have high cell motility: $10\%$ of the cells have an overlap of less than 0.6, furthermore, both data sets have more than 10 cells in the last frame. Data sets with high cell count, more than 100 cells in the last frame, are BF-C2DL-HSC, Fluo-N2DL-HeLa and PhC-C2DL-PSC.

The results on the CTB show, that our approach handles both data sets with high cell motility well, outperforming all other submitted approaches. Also, on data sets with high cell count, our approach performs on at least one of the CTB metrics within the top 2 approaches. Besides that, we perform on three more data sets, Fluo-N2DH-GOWT1,PhC-C2DL-U373, and Fluo-N2DH-SIM+ within the top 3 concerning the overall OP\textsubscript{CTB} metric. Moreover, all nine data sets can be processed within a reasonable time.

\section{Conclusion}
We proposed learning cell segmentation and tracking jointly in a single CNN by learning offsets of cell pixels to their cell centers and a clustering bandwidth. The proposed network architecture does not require any recurrent network parts and the predicted embeddings, offsets and clustering bandwidth, are simple to interpret. To show the performance of EmbedTrack, we evaluated our approach on nine 2D data sets from the Cell Tracking Challenge, which was presented in Table~\ref{tab:ctc_results}. While having a reasonable run time, our approach performs on at least one metric on seven out of nine data sets within the top 3 contestants including three top 1 performances.
Directions of future work are an extension to 3D, training the approach on simulated data using GANs as in~\cite{liu_towards_2021}, or training on sparse labeled data.

\section*{Acknowledgment}
We would like to thank Tim Scherr and Moritz Böhland for proofreading.



\begin{thebibliography}{10}
\providecommand{\url}[1]{#1}
\csname url@samestyle\endcsname
\providecommand{\newblock}{\relax}
\providecommand{\bibinfo}[2]{#2}
\providecommand{\BIBentrySTDinterwordspacing}{\spaceskip=0pt\relax}
\providecommand{\BIBentryALTinterwordstretchfactor}{4}
\providecommand{\BIBentryALTinterwordspacing}{\spaceskip=\fontdimen2\font plus
\BIBentryALTinterwordstretchfactor\fontdimen3\font minus
  \fontdimen4\font\relax}
\providecommand{\BIBforeignlanguage}[2]{{%
\expandafter\ifx\csname l@#1\endcsname\relax
\typeout{** WARNING: IEEEtran.bst: No hyphenation pattern has been}%
\typeout{** loaded for the language `#1'. Using the pattern for}%
\typeout{** the default language instead.}%
\else
\language=\csname l@#1\endcsname
\fi
#2}}
\providecommand{\BIBdecl}{\relax}
\BIBdecl

\bibitem{ulman_objective_2017}
V.~Ulman, M.~Maška, K.~E.~G. Magnusson, O.~Ronneberger, C.~Haubold, N.~Harder,
  P.~Matula, P.~Matula, D.~Svoboda, M.~Radojevic, I.~Smal, K.~Rohr, J.~Jaldén,
  H.~M. Blau, O.~Dzyubachyk, B.~Lelieveldt, P.~Xiao, Y.~Li, S.-Y. Cho, A.~C.
  Dufour, J.-C. Olivo-Marin, C.~C. Reyes-Aldasoro, J.~A. Solis-Lemus,
  R.~Bensch, T.~Brox, J.~Stegmaier, R.~Mikut, S.~Wolf, F.~A. Hamprecht,
  T.~Esteves, P.~Quelhas, {\"O}.~Demirel, L.~Malmström, F.~Jug, P.~Tomancak,
  E.~Meijering, A.~Muñoz-Barrutia, M.~Kozubek, and C.~Ortiz-de Solorzano, ``An
  objective comparison of cell-tracking algorithms,'' \emph{Nature Methods},
  vol.~14, no.~12, pp. 1141--1152, 2017.

\bibitem{stegmaier_fuzzy-based_2017}
J.~Stegmaier and R.~Mikut, ``Fuzzy-based propagation of prior knowledge to
  improve large-scale image analysis pipelines,'' \emph{{PLOS} {ONE}}, vol.~12,
  no.~11, p. e0187535, 2017.

\bibitem{chang_automated_2017}
Y.-H. Chang, H.~Yokota, K.~Abe, C.-T. Tang, and M.-D. Tasi, ``Automated
  detection and tracking of cell clusters in time-lapse fluorescence microscopy
  images,'' \emph{Journal of Medical and Biological Engineering}, vol.~37,
  no.~1, pp. 18--25, 2017.

\bibitem{hirose_spf-celltracker_2018}
O.~Hirose, S.~Kawaguchi, T.~Tokunaga, Y.~Toyoshima, T.~Teramoto, S.~Kuge,
  T.~Ishihara, Y.~Iino, and R.~Yoshida, ``{SPF}-{CellTracker}: Tracking
  multiple cells with strongly-correlated moves using a spatial particle
  filter,'' \emph{{IEEE}/{ACM} Transactions on Computational Biology and
  Bioinformatics}, vol.~15, no.~6, pp. 1822--1831, 2018.

\bibitem{xu_spatial-temporal_2019}
J.~Xu, Y.~Cao, Z.~Zhang, and H.~Hu, ``Spatial-temporal relation networks for
  multi-object tracking,'' in \emph{2019 {IEEE}/{CVF} International Conference
  on Computer Vision ({ICCV})}.\hskip 1em plus 0.5em minus 0.4em\relax {IEEE},
  Oct. 2019, pp. 3987--3997.

\bibitem{hossain_visual_2018}
M.~I. Hossain, A.~K. Gostar, A.~Bab-Hadiashar, and R.~Hoseinnezhad, ``Visual
  mitosis detection and cell tracking using labeled multi-{Bernoulli} filter,''
  in \emph{2018 21st International Conference on Information Fusion
  (FUSION)}.\hskip 1em plus 0.5em minus 0.4em\relax {IEEE}, Jul. 2018, pp.
  1--5.

\bibitem{padfield_coupled_2011}
D.~Padfield, J.~Rittscher, and B.~Roysam, ``Coupled minimum-cost flow cell
  tracking for high-throughput quantitative analysis,'' \emph{Medical Image
  Analysis}, vol.~15, no.~4, pp. 650--668, 2011.

\bibitem{magnusson_global_2015}
K.~E.~G. Magnusson, J.~Jaldén, P.~M. Gilbert, and H.~M. Blau, ``Global linking
  of cell tracks using the {Viterbi} algorithm,'' \emph{{IEEE} Transactions on
  Medical Imaging}, vol.~34, no.~4, pp. 911--929, 2015.

\bibitem{arbelle_probabilistic_2018}
A.~Arbelle, J.~Reyes, J.-Y. Chen, G.~Lahav, and T.~R. Raviv, ``A probabilistic
  approach to joint cell tracking and segmentation in high-throughput
  microscopy videos,'' \emph{Medical Image Analysis}, vol.~47, pp. 140--152,
  2018.

\bibitem{turetken_network_2017}
E.~Türetken, X.~Wang, C.~J. Becker, C.~Haubold, and P.~Fua, ``Network flow
  integer programming to track elliptical cells in time-lapse sequences,''
  \emph{{IEEE} Transactions on Medical Imaging}, vol.~36, no.~4, pp. 942--951,
  2017.

\bibitem{schiegg_graphical_2015}
M.~Schiegg, P.~Hanslovsky, C.~Haubold, U.~Koethe, L.~Hufnagel, and F.~A.
  Hamprecht, ``Graphical model for joint segmentation and tracking of multiple
  dividing cells,'' \emph{Bioinformatics}, vol.~31, no.~6, pp. 948--956, 2015.

\bibitem{akram_cell_2017}
S.~U. Akram, J.~Kannala, L.~Eklund, and J.~Heikkilä, ``Cell tracking via
  proposal generation and selection,'' \emph{{arXiv} preprint}, 2017.

\bibitem{loffler_graph-based_2021}
K.~Löffler, T.~Scherr, and R.~Mikut, ``A graph-based cell tracking algorithm
  with few manually tunable parameters and automated segmentation error
  correction,'' \emph{{PLOS} {ONE}}, vol.~16, no.~9, p. e0249257, 2021.

\bibitem{stringer_cellpose_2021}
C.~Stringer, T.~Wang, M.~Michaelos, and M.~Pachitariu, ``Cellpose: a generalist
  algorithm for cellular segmentation,'' \emph{Nature Methods}, vol.~18, no.~1,
  pp. 100--106, 2021.

\bibitem{cutler_omnipose_2021}
K.~J. Cutler, C.~Stringer, P.~A. Wiggins, and J.~D. Mougous, ``Omnipose: a
  high-precision morphology-independent solution for bacterial cell
  segmentation,'' \emph{bioRxiv preprint}, 2021.

\bibitem{scherr_cell_2020}
T.~Scherr, K.~Löffler, M.~Böhland, and R.~Mikut, ``Cell segmentation and
  tracking using {CNN}-based distance predictions and a graph-based matching
  strategy,'' \emph{{PLOS} {ONE}}, vol.~15, no.~12, p. e0243219, 2020.

\bibitem{weigert_star-convex_2020}
M.~Weigert, U.~Schmidt, R.~Haase, K.~Sugawara, and G.~Myers, ``Star-convex
  polyhedra for 3{D} object detection and segmentation in microscopy,'' in
  \emph{2020 {IEEE} Winter Conference on Applications of Computer Vision
  ({WACV})}, Mar. 2020, pp. 3655--3662.

\bibitem{pal_deep_2021}
S.~K. Pal, A.~Pramanik, J.~Maiti, and P.~Mitra, ``Deep learning in multi-object
  detection and tracking: state of the art,'' \emph{Applied Intelligence},
  vol.~51, no.~9, pp. 6400--6429, 2021.

\bibitem{l_kalake_analysis_2021}
{L. Kalake}, {W. Wan}, and {L. Hou}, ``Analysis based on recent deep learning
  approaches applied in real-time multi-object tracking: A review,''
  \emph{{IEEE} Access}, vol.~9, pp. 32\,650--32\,671, 2021.

\bibitem{ciaparrone_deep_2020}
G.~Ciaparrone, F.~Luque~Sánchez, S.~Tabik, L.~Troiano, R.~Tagliaferri, and
  F.~Herrera, ``Deep learning in video multi-object tracking: A survey,''
  \emph{Neurocomputing}, vol. 381, pp. 61--88, 2020.

\bibitem{payer_segmenting_2019}
C.~Payer, D.~Štern, M.~Feiner, H.~Bischof, and M.~Urschler, ``Segmenting and
  tracking cell instances with cosine embeddings and recurrent hourglass
  networks,'' \emph{Medical Image Analysis}, vol.~57, pp. 106--119, 2019.

\bibitem{zhao_voxelembed_2021}
M.~Zhao, Q.~Liu, A.~Jha, R.~Deng, T.~Yao, A.~Mahadevan-Jansen, M.~J. Tyska,
  B.~A. Millis, and Y.~Huo, ``{VoxelEmbed}: 3{D} instance segmentation and
  tracking with voxel embedding based deep learning,'' in \emph{Machine
  Learning in Medical Imaging}, C.~Lian, X.~Cao, I.~Rekik, X.~Xu, and P.~Yan,
  Eds.\hskip 1em plus 0.5em minus 0.4em\relax Cham: Springer International
  Publishing, 2021, pp. 437--446.

\bibitem{hayashida_mpm_2020}
J.~Hayashida, K.~Nishimura, and R.~Bise, ``{MPM}: Joint representation of
  motion and position map for cell tracking,'' in \emph{2020 {IEEE}/{CVF}
  Conference on Computer Vision and Pattern Recognition ({CVPR})}, Jun. 2020,
  pp. 3822--3831.

\bibitem{he_cell_2017}
T.~He, H.~Mao, J.~Guo, and Z.~Yi, ``Cell tracking using deep neural networks
  with multi-task learning,'' \emph{Image and Vision Computing}, vol.~60, pp.
  142--153, 2017.

\bibitem{shen_cell_2019}
J.~Hayashida and R.~Bise, ``Cell tracking with deep learning for cell detection
  and motion estimation in low-frame-rate,'' in \emph{Medical Image Computing
  and Computer Assisted Intervention – {MICCAI} 2019}.\hskip 1em plus 0.5em
  minus 0.4em\relax Springer International Publishing, 2019, vol. 11764, pp.
  397--405, series Title: Lecture Notes in Computer Science.

\bibitem{wen_3deecelltracker_2021}
C.~Wen, T.~Miura, V.~Voleti, K.~Yamaguchi, M.~Tsutsumi, K.~Yamamoto, K.~Otomo,
  Y.~Fujie, T.~Teramoto, T.~Ishihara, K.~Aoki, T.~Nemoto, E.~M. Hillman, and
  K.~D. Kimura, ``{3DeeCellTracker}, a deep learning-based pipeline for
  segmenting and tracking cells in 3{D} time lapse images,'' \emph{{eLife}},
  vol.~10, p. e59187, 2021.

\bibitem{xie_deep_2021}
Y.~Xie, M.~Liu, S.~Zhou, and Y.~Wang, ``A deep local patch matching network for
  cell tracking in microscopy image sequences without registration,''
  \emph{{IEEE}/{ACM} Transactions on Computational Biology and Bioinformatics},
  pp. 1--1, 2021.

\bibitem{chen_celltrack_2021}
Y.~Chen, Y.~Song, C.~Zhang, F.~Zhang, L.~O'Donnell, W.~Chrzanowski, and W.~Cai,
  ``{CellTrack} {R-CNN}: A novel end-to-end deep neural network for cell
  segmentation and tracking in microscopy images,'' \emph{{arXiv} preprint},
  2021.

\bibitem{sugawara_tracking_2021}
K.~Sugawara, C.~Cevrim, and M.~Averof, ``Tracking cell lineages in 3{D} by
  incremental deep learning,'' \emph{bioRxiv preprint}, 2021.

\bibitem{lugagne_delta_2020}
J.-B. Lugagne, H.~Lin, and M.~J. Dunlop, ``{DeLTA}: Automated cell
  segmentation, tracking, and lineage reconstruction using deep learning,''
  \emph{{PLOS} Computational Biology}, vol.~16, no.~4, p. e1007673, 2020.

\bibitem{ben-haim_graph_2022}
T.~Ben-Haim and T.~Riklin-Raviv, ``Graph neural network for cell tracking in
  microscopy videos,'' \emph{{arXiv} preprint}, 2022.

\bibitem{liu_towards_2021}
Q.~Liu, I.~M. Gaeta, M.~Zhao, R.~Deng, A.~Jha, B.~A. Millis,
  A.~Mahadevan-Jansen, M.~J. Tyska, and Y.~Huo, ``Towards annotation-free
  instance segmentation and tracking with adversarial simulations,''
  \emph{{arXiv} preprint}, 2021.

\bibitem{nishimura_weakly-supervised_2020}
K.~Nishimura, J.~Hayashida, C.~Wang, D.~F.~E. Ker, and R.~Bise,
  ``Weakly-supervised cell tracking via backward-and-forward propagation,'' in
  \emph{Computer Vision – {ECCV} 2020}.\hskip 1em plus 0.5em minus
  0.4em\relax Springer International Publishing, 2020, pp. 104--121.

\bibitem{wang_deep_2020}
J.~Wang, X.~Su, L.~Zhao, and J.~Zhang, ``Deep reinforcement learning for data
  association in cell tracking,'' \emph{Frontiers in Bioengineering and
  Biotechnology}, vol.~8, no. 298, 2020.

\bibitem{neven_instance_2019}
D.~Neven, B.~D. Brabandere, M.~Proesmans, and L.~Van~Gool, ``Instance
  segmentation by jointly optimizing spatial embeddings and clustering
  bandwidth,'' in \emph{2019 {IEEE}/{CVF} Conference on Computer Vision and
  Pattern Recognition ({CVPR})}.\hskip 1em plus 0.5em minus 0.4em\relax {IEEE},
  Jun. 2019, pp. 8829--8837.

\bibitem{romera_erfnet_2018}
E.~Romera, J.~M. Álvarez, L.~M. Bergasa, and R.~Arroyo, ``{ERFNet}: Efficient
  residual factorized {ConvNet} for real-time semantic segmentation,''
  \emph{{IEEE} Transactions on Intelligent Transportation Systems}, vol.~19,
  no.~1, pp. 263--272, 2018.

\bibitem{lalit_embedding-based_2021}
M.~Lalit, P.~Tomancak, and F.~Jug, ``Embedding-based instance segmentation in
  microscopy,'' in \emph{Proceedings of Machine Learning Research}, vol.
  143.\hskip 1em plus 0.5em minus 0.4em\relax {PMLR}, Jul. 2021, pp. 399--415.

\bibitem{maska_benchmark_2014}
M.~Maška, V.~Ulman, D.~Svoboda, P.~Matula, P.~Matula, C.~Ederra, A.~Urbiola,
  T.~España, S.~Venkatesan, D.~M. Balak, P.~Karas, T.~Bolcková,
  M.~Štreitová, C.~Carthel, S.~Coraluppi, N.~Harder, K.~Rohr, K.~E.~G.
  Magnusson, J.~Jaldén, H.~M. Blau, O.~Dzyubachyk, P.~Křížek, G.~M. Hagen,
  D.~Pastor-Escuredo, D.~Jimenez-Carretero, M.~J. Ledesma-Carbayo,
  A.~Muñoz-Barrutia, E.~Meijering, M.~Kozubek, and C.~Ortiz-de Solorzano, ``A
  benchmark for comparison of cell tracking algorithms,''
  \emph{Bioinformatics}, vol.~30, no.~11, pp. 1609--1617, 2014.

\bibitem{berman_lovasz-softmax_2018}
M.~Berman, A.~R. Triki, and M.~B. Blaschko, ``The {Lovász}-softmax loss: A
  tractable surrogate for the optimization of the intersection-over-union
  measure in neural networks,'' in \emph{Proceedings of the {IEEE} Conference
  on Computer Vision and Pattern Recognition ({CVPR})}, Jul. 2018, pp.
  4413--4421.

\bibitem{yu_learning_2015}
J.~Yu and M.~Blaschko, ``Learning submodular losses with the {L}ovász hinge,''
  in \emph{Proceedings of the 32nd International Conference on Machine
  Learning}, ser. Proceedings of Machine Learning Research, vol.~37, Jul. 2015,
  pp. 1623--1631.

\bibitem{matula_cell_2015}
P.~Matula, M.~Maška, D.~V. Sorokin, P.~Matula, C.~Ortiz-de Solórzano, and
  M.~Kozubek, ``Cell tracking accuracy measurement based on comparison of
  acyclic oriented graphs,'' \emph{{PLOS} {ONE}}, vol.~10, no.~12, p. e0144959,
  2015.

\bibitem{kingma_adam_2017}
D.~P. Kingma and J.~Ba, ``Adam: A method for stochastic optimization,''
  \emph{{arXiv} preprint}, 2014.

\end{thebibliography}

\end{document}